\documentclass[a4paper, 11pt]{article}
\usepackage[utf8]{inputenc}
\usepackage[english]{babel}
\usepackage[T1]{fontenc}

\usepackage{arxiv}

\usepackage{physics}
\usepackage{amsmath}
\usepackage[colorlinks=true, allcolors=blue]{hyperref}
\usepackage{amssymb}
\usepackage{mathtools}
\usepackage{ mathrsfs }
\usepackage{bbm}
\usepackage{bm}
\usepackage{soul}

\usepackage{xcolor}
\definecolor{gblue}{rgb}{0.0,0.7,0.6}

\definecolor{orange}{rgb}{1.0,0.5,0.0}

\usepackage[backend=biber, 
            natbib=true,
            style=numeric-comp,
            sorting=none]{biblatex}

\newcommand{\eg}{{\it e.g.}}
\newcommand{\ie}{{\it i.e.}}

\newcommand{\Win}{\mathbf{W_{\rm in}}}
\newcommand{\Wout}{\mathbf{W_{\rm out}}}

\def\r{\vb r}

\def\A{\vb A}

\def\u{\vb u}

\newcommand{\be}{\begin{equation}}
\newcommand{\bea}{\begin{eqnarray}}
\newcommand{\ee}{\end{equation}}
\newcommand{\eea}{\end{eqnarray}}

\def\A{\mbox{\bf A}}

\def\r{\mbox{{\bf r}}}

\def\u{\mbox{$\bf{u}$}}

\title{Constraining Chaos: Enforcing dynamical invariants in the training of recurrent neural networks}
\author{Jason A. Platt \thanks{Department of Physics, University of California San Diego,
La Jolla, CA 92093} \and Stephen G. Penny \thanks{Cooperative Institute for Research in Environmental Sciences,
University of Colorado Boulder,
Boulder, CO 80309,
Physical Sciences Laboratory,
National Oceanic and Atmospheric Administration,
Boulder, CO 80305}
\and Timothy A. Smith \footnotemark[2]
\and Tse-Chun Chen \footnotemark[2]
\and Henry D. I. Abarbanel \footnotemark[1]$^{\ }$ 
\thanks{Marine Physical Laboratory,
Scripps Institution of Oceanography,
University of California San Diego,
9500 Gilman Drive,
La Jolla, CA 92093}}
\date{\today}

\begin{document}

\maketitle
\begin{abstract}
Drawing on ergodic theory, we introduce a novel training method for machine learning based forecasting methods for chaotic dynamical systems.  The training  enforces dynamical invariants---such as the Lyapunov exponent spectrum and fractal dimension---in the systems of interest, enabling longer and more stable forecasts when operating with limited data.  The technique is demonstrated in detail using the recurrent neural network architecture of reservoir computing. Results are given for the Lorenz 1996 chaotic dynamical system and a spectral quasi-geostrophic model, both typical test cases for numerical weather prediction.
\end{abstract}

\section{Introduction}

Predicting the future trajectory of a dynamical system---a time series whose evolution is governed by a set of differential equations---is crucial in fields such as weather prediction, economics, chemistry, physics and many others \cite{pred_future, strogatz00}.  A prediction can be generated by deriving the governing equations of motion (EOM) for the system and integrating forward in time, perhaps with data being used to determine the value of particular constants or the initial conditions. Machine learning (ML), on the other hand, allows the construction of a forecast purely from observational data in lieu of a physical model. When the EOM are expensive to evaluate numerically, ML can be used to construct a surrogate model; such models can be integrated into data assimilation \cite{kalnay2002} algorithms---such as the Kalman filter \cite{Kalman60, mandel09}---a typical use case when data are noisy and the model imperfect, such as in numerical weather prediction \cite{penny22}.

The inclusion of physical knowledge---EOM, conservation laws and dynamical invariants---into ML algorithms has been a topic of ongoing interest \cite{Karniadakis21, Seungwoong21, Alet21, Ziming21, Beucler21, Chen20, Greydanus19, Raissi19, yang21}. Enforcing these laws effectively reduces the searchable parameter space for a workable model, decreasing the training time and increasing the accuracy of the resulting models. An ML model trained without knowledge of invariants may fail to generalize and can produce solutions that violate fundamental constraints on the physical system \cite{Azizi21}. Many of the examples cited above involve conservation of quantities based on the symmetry of the equations of motion, such as conservation of energy and momentum \cite{Greydanus19}, or the inclusion of previously derived differential equations \cite{Beucler21} as components of the ML training.  ``Physics informed'' neural networks \cite{Raissi19, yang21, Doan20, Doan21, Racca21} add the known or partially known differential equations as a soft constraint in the loss function of the neural network, but conservation laws are not necessarily enforced and the equations need to be known.  

Many physical dynamical systems of interest are dissipative---\eg, any dynamical system containing friction---meaning they exchange energy and mass with the surrounding environment \cite{Goldstein2001}. High dimensional dissipative systems are very likely to exhibit chaos \cite{Ispolatov15}---making them extremely sensitive to initial conditions. Enforcing conservation of quantities such as momentum, mass, or energy \cite{Greydanus19, zanna_data-driven_2020, beucler_towards_2020, Beucler21} for dissipative systems in isolation may not be sufficient for generalization due to the exchange of energy/momentum at the boundaries. Problems concerning chaotic dynamics, such as weather forecasting, exhibit fractal phase space trajectories that make it difficult to write down analytic constraints \cite{Beucler21}.   


With the goal of enforcing dynamical invariants, we suggest an alternative cost function for dissipative systems based on ergodicity, rather than symmetry. This has broad implications for time series prediction of dynamical systems. 
After formulating the invariants, we give a recurrent neural network (RNN) \cite{Goodfellow16} example applied to the Lorenz 1996 system \cite{Lorenz96} and quasi-geostrophic dynamics \cite{Reinhold82} where we add soft constraints into the training of the network in order to ensure that these quantities are conserved.

\section{Deriving Dynamical Invariants}

Ergodicity is a property of the evolution of a dynamical system.  A system exhibiting ergodicity, called ergodic, is one in which the trajectories of that system will eventually visit the entire available phase space \cite{Datseris2022}, with time spent in each part proportional to its volume.  In general, the available phase space is a subset of the entire phase space volume. For instance a Hamiltonian system will only visit the hypersurface with constant energy \cite{Goldstein2001}. Ergodicity implies that time averages over the system trajectories can be replaced with spatial averages
\begin{equation}
    \lim_{t_f \to \infty} \frac{1}{t_f} \int_0^{t_f} g(F^t(\u_0))dt = \int_{\u \in B} g(\u) \rho_B(\u) du \qquad \forall \u_0 \in B \label{eq: ergodic_thm}
\end{equation}
for an arbitrary function $g$, where $F^t$ is the application of the flow of the dynamical system of $t$ iterations, and $\rho_B(\u)$ defines an invariant density over the finite set $B$ \cite{Datseris2022}. The invariant density gives an intuitive measure of how often a trajectory visits each part of $B$. The invariant density defines the invariant measure \cite{Datseris2022}
\begin{equation}
    \mu(B \subset R) = \int_B \rho_B(u)du. \label{eq: invariant_measure}
\end{equation}

\begin{figure}[!htpb]
    \centering
    \includegraphics[width=0.85\textwidth]{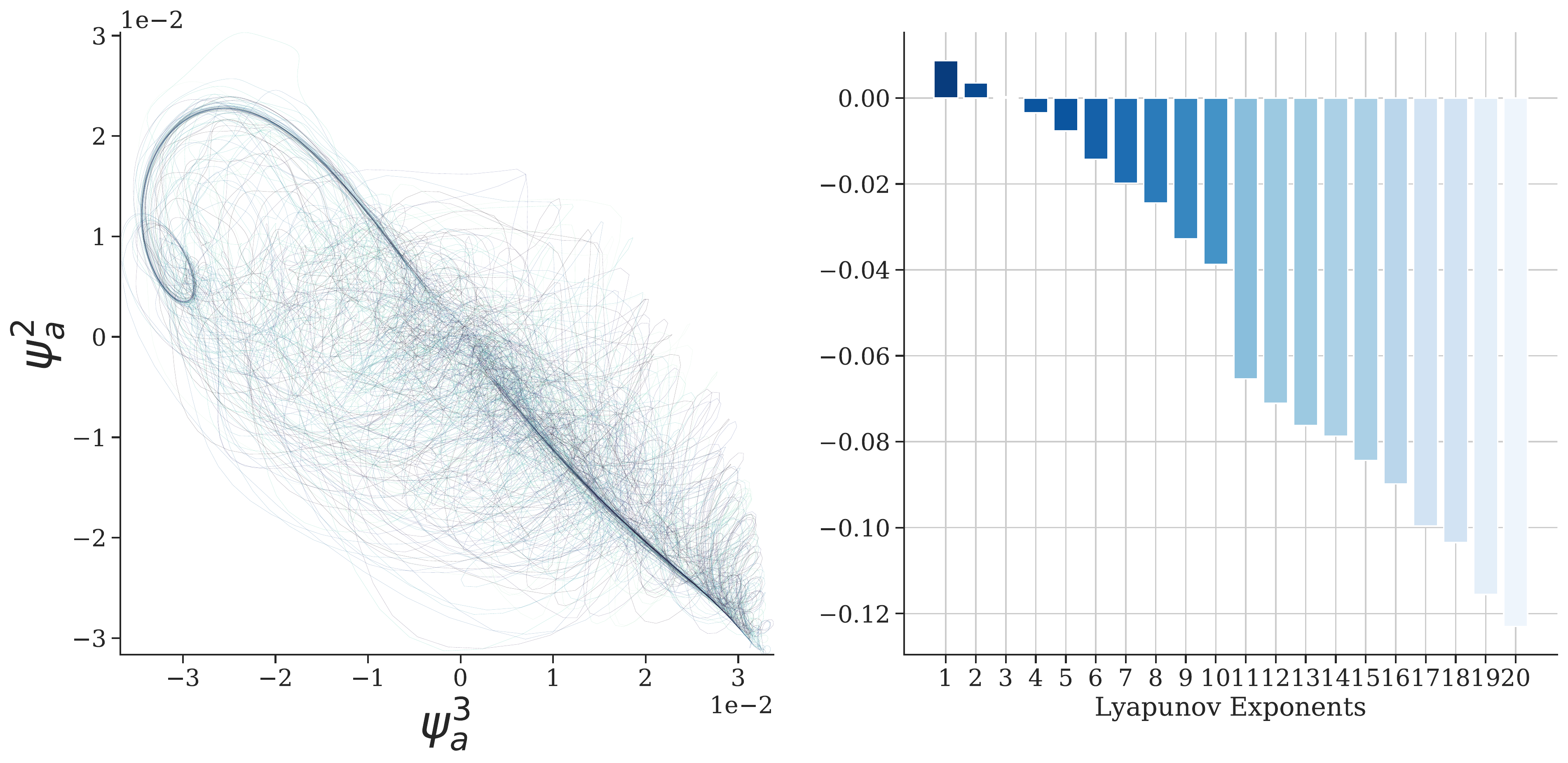}
    \caption{left) 2D slice through the strange attractor of the spectral QG model reproduced from Fig.(3) in \cite{Reinhold82} using the implementation in \cite{qgs}.  An attractor is a hypersurface that draws in nearby trajectories of the system, such that the system will eventually be constrained to stay on the manifold. The motion of the dynamical system on a strange attractor is chaotic with extreme sensitivity to initial conditions. Strange attractors can be analyzed through the invariant measure that describes how often the system visits each part of the attractor \cite{abar96, Eckmann85}. Related quantities such as the fractal dimension and the Lyapunov exponents \cite{Eckmann85} are globally invariant through any smooth change of coordinates and thus are natural invariant quantities for chaotic systems. right) The Lyapunov exponents of the spectral QG model.  There are two positive exponents making the system chaotic.}
    \label{fig: qg_attractor}
\end{figure}

$B$ will often consist of exotic geometries such as quasi-periodic orbits and strange attractors \cite{Goldstein2001}. The strange attractor is a hypersurface that contains the available subspace for a chaotic dynamical system---see Fig.(\ref{fig: qg_attractor}). Deterministic chaotic systems are of importance to a vast array of applications such as in numerical weather prediction \cite{kalnay2002}, chemical mixing \cite{Ottino04,Grigoriev06}, optics \cite{Arecchi90}, robotics \cite{Zang16} and many other fields.


Despite being deterministic, the precise long term prediction of chaotic systems is impossible due to the exponential growth of errors, as quantified by the system's Lyapunov spectrum. The Lyapunov spectrum, composed of a system's Lyapunov exponents (LEs), characterizes a dynamical system \cite{lyapunov, abar96} by giving a quantitative measure of how a volume of phase space stretches or shrinks over time.


For the prediction of chaotic systems, we suggest that although short term predictions will inevitably diverge, long term prediction of any system must preserve the invariants of motion characterized by the invariant measure $\mu_B$ Eq.(\ref{eq: invariant_measure}). Furthermore, enforcing such invariants could help to generalize the training of neural networks designed to emulate dissipative chaotic systems, in much the same way that conservation of energy and momentum has for conservative systems. While any function $g(\u)$ integrated with the invariant density is a constant---as seen in Eq.(\ref{eq: ergodic_thm})---by the multiplicative ergodic theorem \cite{ose68} the LEs and the fractal dimension are both invariant under smooth coordinate transformations and have algorithms that make them feasible to compute from observed data \cite{abar96}. 

In the next sections we provide a concrete example using the fractal dimension and LEs as invariants that must be enforced when training a neural network. We use an RNN based on the reservoir computer (RC) architecture \cite{Maass02,Jaeger01,Lukoševičius12}. We impose a loss function that takes into account the preservation of the LEs and fractal dimension and detail the benefits of doing so. We stress that the concept is not limited to RC models and can in fact apply to any neural network architecture.

\section{Recurrent Neural Networks and Reservoir Computing}
An RNN is a network composed of nonlinear elements that are connected in such a way as to enable self excitation \cite{elman1990}. Therefore, given a state of the network $\r(t-1)$, the next state
\begin{equation}
    \r(t) = F_r(\r(t-1), \u(t-1), \theta) \label{eq: abstract_driven_rnn}
\end{equation}
is a function of the input $\u$, the RNN equations $F_r$ and the internal weights $\theta$. The label over the input data $t\in \mathbb{Z}$---conveniently called time---gives the natural order and allows the analysis of the RNN as a dynamical map. $\r(t)$ can then be decoded by a function $W_{\rm out}(\r(t)) = \hat \u$.  $W_{\rm out}$ is trained so that $\hat \u$ is as close to the target output as possible \cite{Goodfellow16}. In time series prediction tasks $\hat \u \sim \u(t)$ so that the driven system Eq.(\ref{eq: abstract_driven_rnn}) can become autonomous (with no external input)
\begin{equation}
    \r(t) = F_r(\r(t-1), W_{\rm out}(\r(t-1)), \theta) \label{eq: abstract_auto_rnn}
\end{equation}
and predict the future of the dynamical system.


Reservoir computing (RC) \cite{Jaeger01, Jaeger02, Jaeger04, Maass02, luk09, Jaeger12} is a simplified form of RNN for which only the large scale parameters of the network are varied with the detailed weights selected from probability distributions. For an RC with $\tanh$ units at the nodes the RNN equations become \cite{platt22}
\begin{equation}
    \r(t) = \alpha \tanh(\A \r(t-1) + \Win \u(t-1) + \sigma_b) + (1-\alpha) \r(t-1).
\end{equation}
The elements of the $N \times N$ adjacency matrix $\A$ are fixed \ie, not trained---in contrast to other RNN architectures---with only its overall properties chosen such as the size $N$, density $\rho_A$ and spectral radius $\rho_{SR}$. $\Win \in \mathbb{R}^{N \times D}$ maps the input into the high dimensional reservoir space $\u \in \mathbb{R}^D \to \Win \u \in \mathbb{R}^N$; the elements of $\Win$ are chosen between $\mqty[-\sigma, \sigma]$. $\alpha$ is the leak rate and is related to the time constant of the RC \cite{Lukoševičius12}. $\sigma_b$ is an input bias governing the fixed point of the RC and the strength of the $\tanh$ nonlinearity. See \cite{Lukoševičius12} for detailed explanations of the architecture and parameter choices.

Training the RC includes training the function $W_{\rm out}$---often taken to be a matrix $\Wout$ and trained through linear regression---as well as finding the correct parameters $N,\ \rho_A,\ \rho_{SR},\ \sigma,\ \sigma_b$. In \cite{Griffith19, penny22, platt22} it is shown how to train the RC network through a two step training procedure that takes into account both the one step prediction accuracy as well as the long term forecast skill.

\begin{figure}
    \centering
    \includegraphics[width=0.8\textwidth]{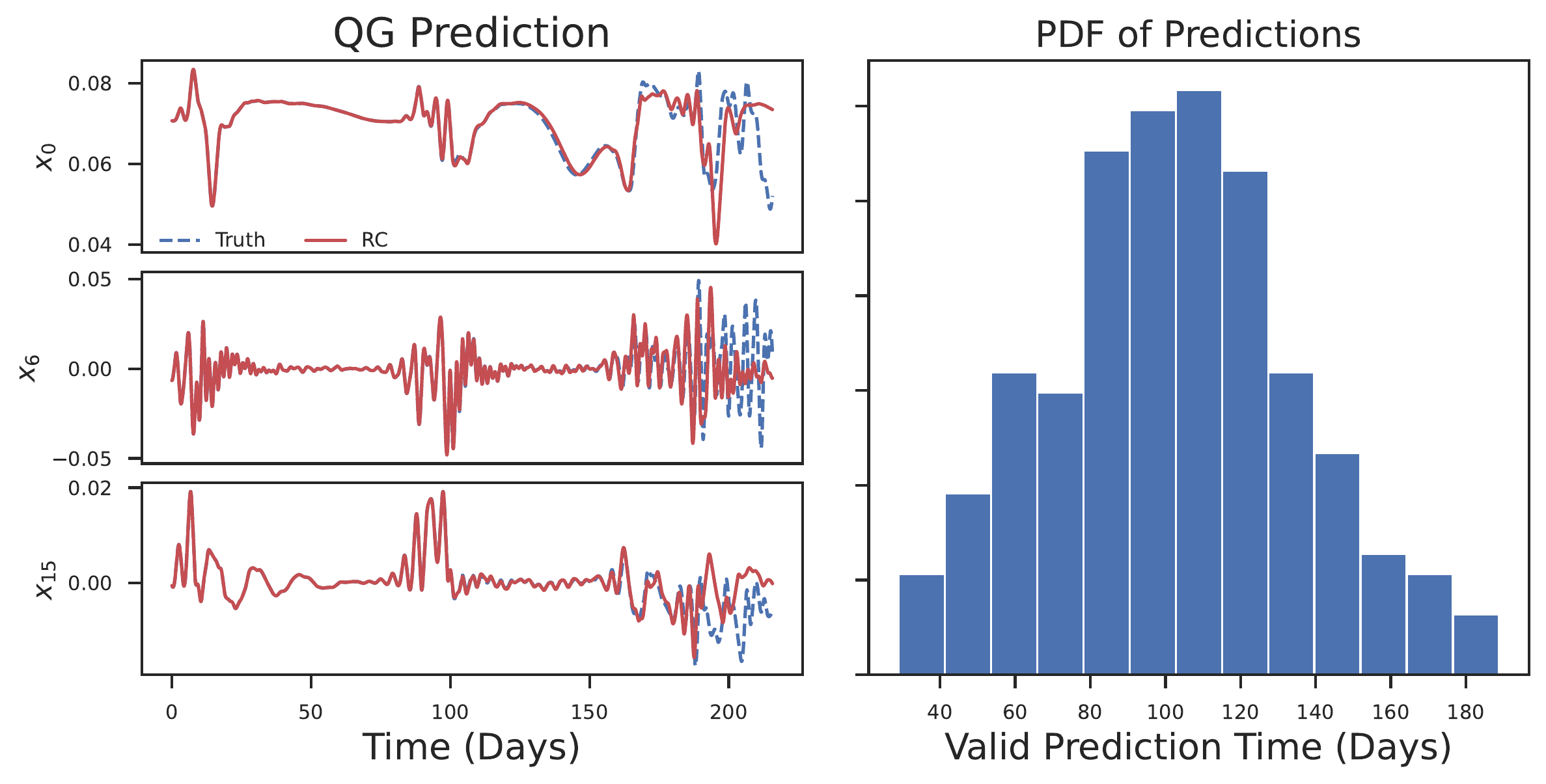}
    \caption{Example prediction and the probability distribution function of the valid prediction time (VPT) for 200 initial conditions over the 20 dimensional QG system described in section \ref{sec: QGS}.  The valid prediction time \cite{Vlachas20, penny22} is calculated as the time when the root mean square error $\rm RMSE(t) = \sqrt{\frac{1}{D} \sum_{i=1}^D \Big [\frac{u_i^f(t) - u_i(t)}{\sigma_i} \Big]^2}$ exceeds a certain value $\epsilon$, in this case $\epsilon = 0.3$ approximately in line with \cite{platt22, Vlachas20, penny22}. $D$ is the system dimension, $\sigma$ is the long term standard deviation of the time series and $u^f$ is the RC forecast.  The largest LE $1/\lambda_1 \sim 12 \text{ days}$ gives the natural time scale for error growth of the system and thus can be used as a measurement for the predictive skill.  The RC returns fantastic prediction times for this low resolution model.  The size of the RC is N=1500 and 100,000 training steps were provided with a $\Delta t = 80 min$ giving about 30 years of data, we consider this the ``data rich'' case.}
    \label{fig: QG_pred}
\end{figure}

RC has been shown to be extremely successful in time series prediction tasks. Its simple form allows the easy computation of the Jacobian and other quantities that can help in a dynamical systems analysis of the RNN. In Platt et al.\cite{platt21} the authors showed that the RC, when well trained, can reproduce invariants of the motion such as the LEs and fractal dimension and that the reproduction of these quantities maximized the prediction time and ensured the stability of the predictions. The training procedure in those previous works does not enforce these invariants explicitly.  The hope in those previous works is that the RC is both capable of reproducing these quantities and that the loss function of short and long term forecasts guides the RC towards these values by proxy. Here we reformulate the training to take into account the invariant quantities.

\section{Enforcing Invariants}
\begin{figure}[!htpb]
    \centering
    \includegraphics[width=0.8\textwidth]{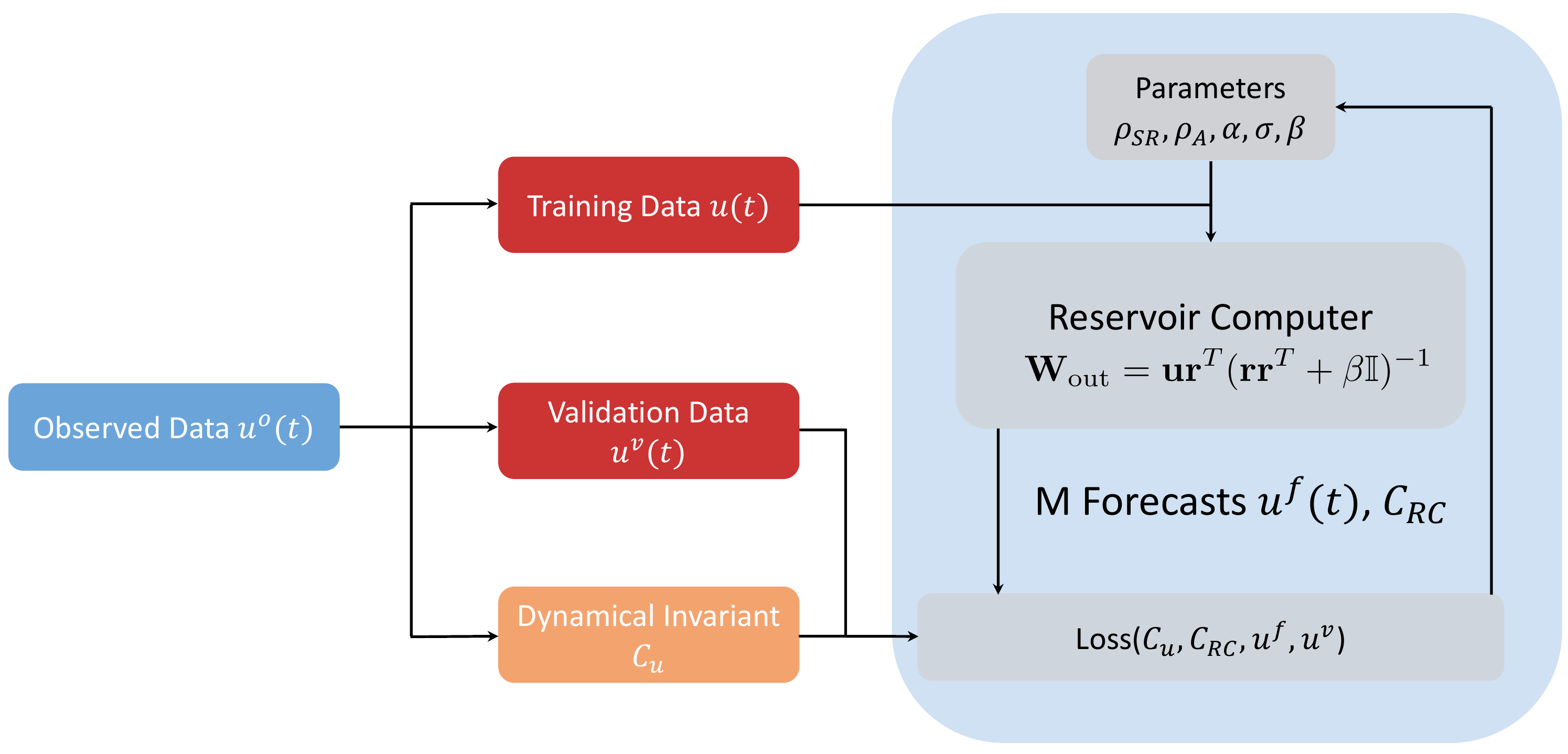}
    \caption{Parameter optimization of a reservoir computer showing the introduction of dynamical invariants into the routine.  The observed data is split into training, validation and testing sets with the invariants calculated from the data \cite{abar96}.  These quantities can then be incorporated into the loss function to improve the overall training of the reservoir computer.  A general discussion of the training strategy is found in \cite{platt22}.}
    \label{fig: global optimization}
\end{figure}

The training of an RC is determined by the training data and the selection of the parameters governing the global properties of the RC: $N,\ \rho_A,\ \rho_{SR},\ \sigma,\ \sigma_b$ and a regularization coefficient $\beta$. Once these quantities are chosen and the weights instantiated, then $\Wout$ is given 
\begin{equation*}
    \Wout = \u \r^T(\r\r^T + \beta \mathbb{I})^{-1}
\end{equation*}
where $\u$ is the $D \times T$ matrix of input data, $T$ is the number of time steps and $\r$ is the $N \times T$ matrix of reservoir states \cite{platt22}.  
For an RC we can add into the selection of these parameters knowledge of the global invariants of the system $\u$. Therefore we construct a loss function
\begin{equation}
        \rm Loss = \epsilon_1\norm{C_{u} - C_{RC}}^2 + \epsilon_2 \sum_{k=1}^M \sum_{t=t_i}^{t_f} \norm{\u_k^f(t) - \u_k(t)}^2 \exp{-\frac{t - t_i}{t_f - t_i}}; t\in \mathbb{Z} \label{eq: global_loss}
\end{equation}
that can be minimized over the RC parameters and with $\epsilon_x$ hyperparameters. The selection of the parameters leads to the matrix $\Wout$ based on training data $\u_{\rm train}$. Platt et al. \cite{platt22} generated a number of long term forecasts $\u^f(t)$ and compared them to the data $\u$; with enough data this procedure often leads to a model that reproduces the correct dynamical invariants. Without the explicit enforcement of these invariants, however, the model can fail to capture the dynamics---particularly for high dimensional systems and in cases where the number of trajectories $M$ is constrained. Here we add the dynamical invariants ($C_x$) as a constraint in order to directly train for generalizability, similar to \cite{beucler_enforcing_2021}. This scheme is illustrated in Fig.(\ref{fig: global optimization}). The global optimization routine used to minimize the cost function was the covariance matrix adaption evolution strategy (CMA-ES) \cite{Hansen03}.

We show the Lyapunov exponents and the fractal dimension as examples of dynamical invariants in order to demonstrate the technique. With the equations of motion, such as Eq.(\ref{eq: abstract_auto_rnn}) for the RC, it is quite simple to calculate these quantities using well known and efficient algorithms \cite{Eckmann85}. When training directly from data---without knowledge of the underlying system---we may not know the equations of motion so these quantities must be estimated. The largest LE can often be approximated from time series data \cite{abar96, Rosenstein93, kantz94} and the fractal dimension can be calculated using various techniques \cite{abar96, Theiler90}. A calculation of the full LE spectrum is more difficult. Use of other dynamical invariants derived from the invariant measure Eq.(\ref{eq: invariant_measure}) are also possible, for instance the energy density spectrum of a fluid dynamical system as a function of wavenumber.

\section{Results}
\subsection{Lorenz 1996}
\begin{figure}[!htpb]
    \centering
    \includegraphics[width=0.6\textwidth]{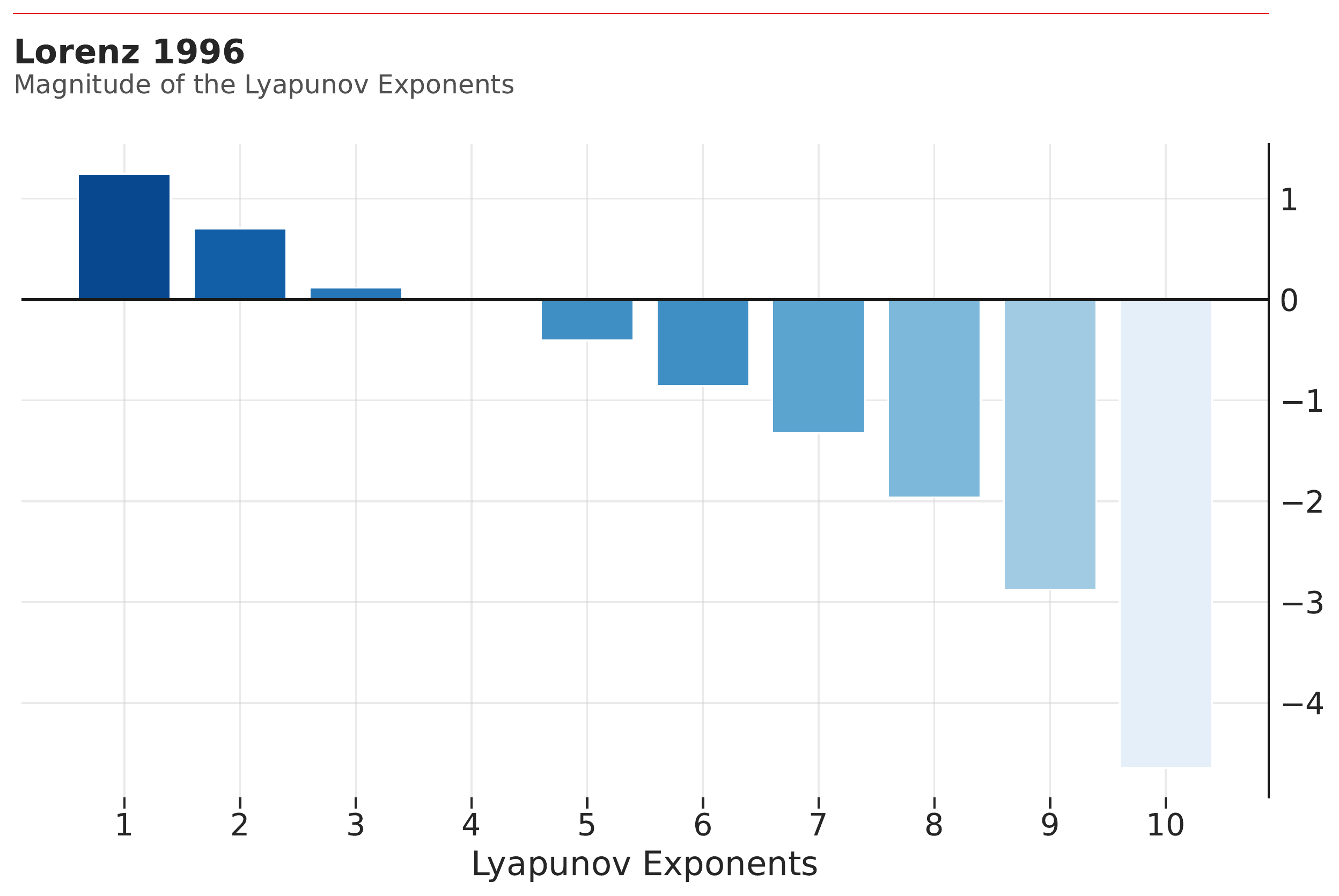}
    \caption{Lyapunov exponents of the 10D Lorenz 1996 system with F=8.  There are three positive LEs, a single zero exponent, and the rest negative.}
    \label{fig: LE_l96}
\end{figure}

Our first test case for the RC is the Lorenz 1996 system (L96), a standard testbed for data assimilation applications in numerical weather prediction. L96 describes the evolution of a scalar quantity over a number of sites scattered uniformly over a periodic spatial lattice of constant latitude with quantities approximating advection and diffusion
\begin{equation}
    \dv{u_k}{t} = -u_{k-1}(u_{k-2} - u_{k+1})-u_k+F. \label{eq: lor96}
\end{equation}
In this case we take the number of sites to be $D=10$ and forcing $F=8$ with the purpose of making the system hyperchaotic, with three positive Lyapunov exponents Fig.(\ref{fig: LE_l96}).

\begin{figure}[!htpb]
    \centering
    \includegraphics[width=0.5\textwidth]{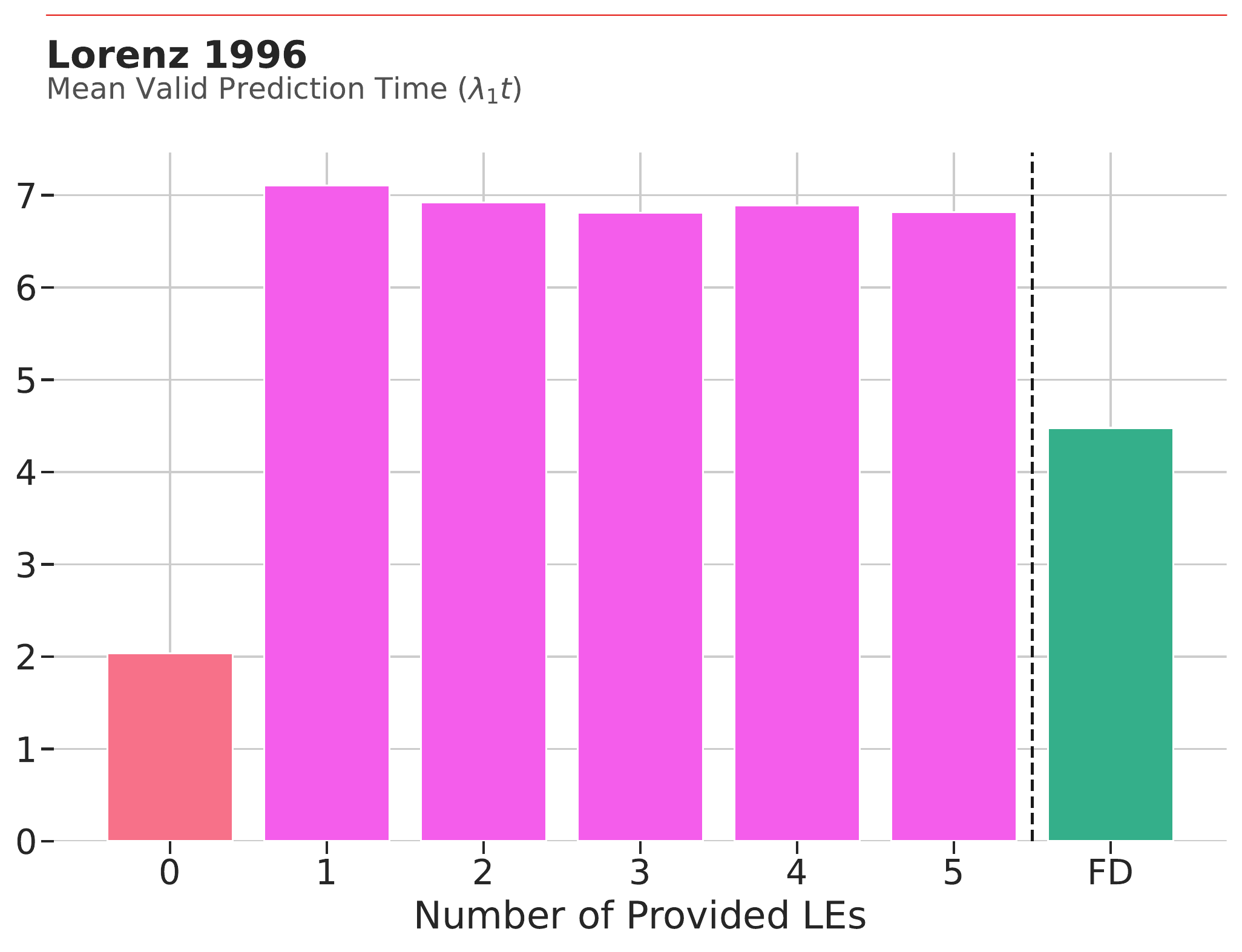}
    \caption{The RC is initialized 10 times, providing k LEs and M=7 long term forecasts Eq\eqref{eq: global_loss}; we report on the average of the distribution of predictions for the 10 dimensional Lorenz 1996 system \cite{Lorenz96} Eq\eqref{eq: lor96}.  There are 10 total LEs---3 positive, shown in Fig.(\ref{fig: lor96_LE}). When 0 LEs are provided the RC has no global information and the prediction time is poor. Giving the largest LE is sufficient for generalizable predictions and this quantity is quite easily calculated from numerical data \cite{abar96}. The last column shows the result for providing the fractal dimension as another example of an invariant quantity. The size of the RC is $N=400$ and the number of test initial conditions was 1000.  The forecast time is given as an average over the 1000 trajectories and then scaled by the largest Lyapunov exponent to give the prediction in terms of the number of Lyapunov timescales.}
    \label{fig: lor96_LE}
\end{figure}

The results for $C_{RC} = LEs$  Eq.(\ref{eq: global_loss}) are shown in Fig.(\ref{fig: lor96_LE}). When no global information is given to the RC then it can fail to generalize when presented with unseen input. Simply providing the largest LE to the RC during training enables the neural networks to
\begin{enumerate}
    \item generalize to unseen data so that there are good predictions over the entire range of possible initial conditions
    \item reconstruct the attractor as in \cite{Lu18} and \cite{platt21} with the prediction giving the correct ergodic properties of the data even after the prediction necessarily diverges from the ground truth.
\end{enumerate}
In this case providing the largest exponent was enough to improve the predictions, with no further gains coming from providing the smaller exponents. This could perhaps be due to the parameter space being constrained enough for those exponents to be matched by the RNN even though they are not directly given.  

The second invariant given is the fractal dimension of the data calculated through the Kaplan-Yorke formulation

\begin{equation}
    \text{Dimension} = \alpha+\frac{\sum_{i=1}^\alpha  \lambda_i}{|\lambda_{\alpha+1}|}, \label{eq: KY}
\end{equation}
with $D$ the system dimension, $\lambda$ the ordered Lyapunov exponents and $\alpha$ the smallest index where the sum of the LE doesn't cross zero \cite{abar96}.  There are alternate definitions and methods for calculating the fractal dimension \cite{Theiler90}.

Providing the fractal dimension as an invariant has a similar effect to providing the LEs by raising the mean valid prediction time from $\sim 2 \rightarrow 4$. The fractal dimension may not, however, be unique to a particular set of data while the LE spectrum has a much greater chance of constraining the shape of the resulting strange attractor. Therefore we see that there is not as much improvement in the forecast when providing the fractal dimension compared to the LE spectrum.
\subsection{Synoptic Scale Atmospheric Model} \label{sec: QGS}
For more complex higher-dimensional dynamical systems, the Lyapunov spectrum or Kaplan-Yorke dimension are quite difficult, if not impossible, to calculate. However, our previous results showed that capturing the leading Lyapunov exponent (LLE) enhanced prediction skill greatly, and even with complex models this quantity can be estimated more readily either from data \cite{Rosenstein93, kantz94} or from a model \cite{Geist90, Benettin76}. We therefore explore the value of representing the LLE in the more complex case of quasi-geostrophic (QG) dynamics \cite{charney48}, which assume that large-scale atmospheric disturbances are governed by the conservation of potential temperature and absolute potential vorticity, while the horizontal velocity is quasi-geostrophic and the pressure quasi-hydrostatic. Numerical models based on the QG approximation were a precursor to larger scale primitive equation models used for global numerical weather prediction \cite{kalnay2002} and is frequently used in data assimilation studies targeting the atmosphere and ocean \cite{Yang09, Swanson98}. Here, we consider the two-layer baroclinic model of Charney and Strauss (1980) \cite{CharneyStraus80} used to study the planetary-scale motions of a thermally driven atmosphere in the presence of topography. We further incorporate the adaption of Reinhold and Pierrehumbert (1982) \cite{Reinhold82} to include an additional wave in the zonal direction making it highly baroclinically unstable. We use the implementation of \cite{qgs}, which provides a truncated 2-layer QG atmospheric model on a mid-latitude $\beta-\text{plane}$ frictionally coupled to a mountain and a valley with a dimension of 20 in the spectral space of the model.

For the atmospheric streamfunctions $\psi_a^1/\psi_a^3$ at heights 250/750 hPa and the vertical velocity $\omega = \dv{p}{t}$, the equations of motion are derived to be
\begin{align}
    &\pdv{t}\overbrace{\mqty(\nabla^2 \psi_a^1)}^{\rm vorticity} + \overbrace{J(\psi_a^1, \nabla^2 \psi_a^1)}^{\text{horizontal advection}} + \overbrace{\beta \pdv{\psi_a^1}{x}}^{\beta-\text{plane Coriolis force}} = -\overbrace{k'_d \nabla^2(\psi_a^1 - \psi_a^3)}^{\rm friction} + \overbrace{\frac{f_0}{\Delta p}\omega}^{\text{vertical stretching}} \\
    &\pdv{t}\mqty(\nabla^2 \psi_a^3) + J(\psi_a^3, \nabla^2 \psi_a^3)+ J(\psi_a^3, f_0 h/H_a) + \beta \pdv{\psi_a^3}{x} = k'_d \nabla^2(\psi_a^1 - \psi_a^3) -k_d \nabla^2 \psi_a^3 + \frac{f_0}{\Delta p}\omega
\end{align}
with $\nabla = \pdv{x} \hat x+\pdv{y} \hat y$, $k'_d$ the friction between the layers, $k_d$ the friction between the atmosphere and the ground, $h/H_a$ the ratio of ground height to the characteristic depth of the atmospheric layer, $\Delta p = 500$ hPa the pressure differential between the layers and $J(g_1, g_2) = \pdv{g_1}{x}\pdv{g_2}{y} - \pdv{g_1}{y}\pdv{g_2}{x}$ the Jacobian.  More details are given in \cite{Reinhold82, qgs}.

After integrating the model forward in time we  ask if an RC model is capable of predicting the dynamics given a significant amount of training data.  An example forecast and distribution of predictions  is shown in Fig.(\ref{fig: QG_pred}) where the RC successfully predicts the synoptic scale atmospheric dynamics for a number of months.  Such significant predictive power on a low resolution QG model is an interesting result in and of itself, showcasing the ability of RNNs to resolve more realistic atmospheric dynamics.

\begin{figure}[!htpb]
    \centering
    \includegraphics[width=0.6\textwidth]{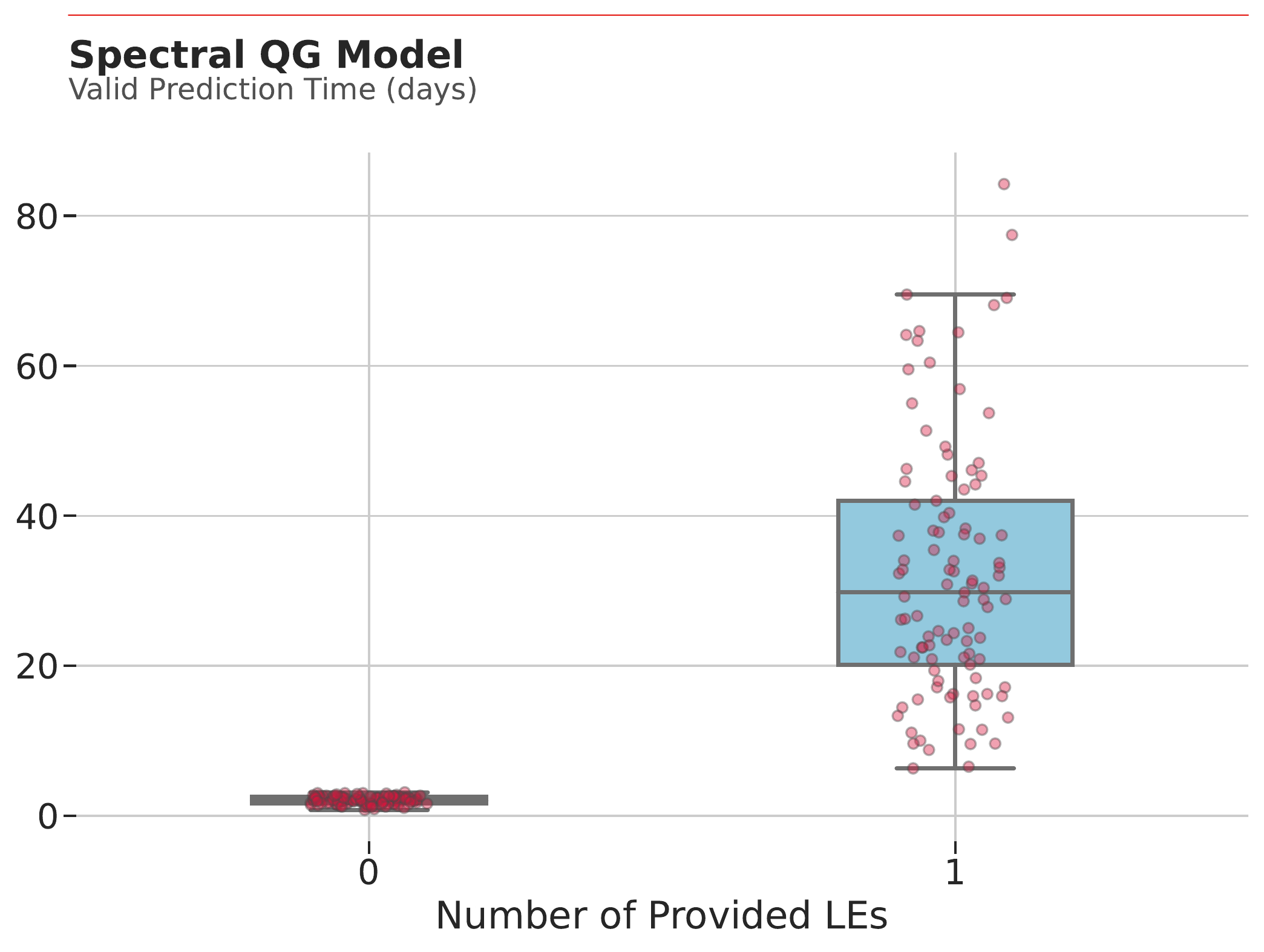}
    \caption{Prediction time in days across 100 predictions for the QG model with an RC trained with and without input of LEs and with $t_f=1$ in the optimization routine Eq.(\ref{eq: global_loss}).  The size of the reservoir is $N=500$ and 50000 training steps are provided with $\Delta t = 80min$. Without the provided exponents in this low information regime, unlike in the data rich example in Fig(\ref{fig: QG_pred}), the RC model is unable to correctly infer the correct dynamical invariants and fails to generalize.}
    \label{fig: QG_prediction_time}
\end{figure}

When we reduce the amount of data and set $M = 1$---the limited data case in Eq. (\ref{eq: global_loss}) with a single  forecast as part of the training loss---the reservoir loses its predictive power.  Adding in the information contained in the LLE, which is $\sim 0.01$, enables the model to recover a large amount of predictive capability. The calculated LLE of the RC with no provided exponents is 0.14, compared to $\sim 0.01$ when it is provided.  The mismatch between the LEs of the two systems is a clear indication that synchronization between the two is not achieved \cite{platt21}. The impact of the added information provided via the LLE is clear in Figure \ref{fig: QG_prediction_time}, where the average VPT has extended from only a few days to multiple weeks.

\section{Discussion and Conclusion}
Chaotic dynamical systems are difficult to predict due to their sensitivity to initial conditions \cite{Lorenz63}. Better understanding and accounting for dynamical uncertainties has, however, allowed fields like numerical weather prediction to provide useful and continually improving forecasts \cite{kalnay03}.  Previous works (e.g. \cite{Greydanus19, beucler_enforcing_2021}) proposed that including conserved quantities such as energy/momentum may help to improve the application of neural networks to physical systems. However, the introduction of the proposed conserved quantities is not generally applicable to dissipative chaotic dynamical systems. Thus, we instead considered dynamical invariants based on the invariant measure.

We provided a concrete example using quantities derived from the invariant measure, such as the Lyapunov exponents and the fractal dimension, to train a particular RNN architecture called reservoir computing. Previous RC training algorithms used long-term forecasts initialized from many different initial conditions in order to improve generalizability \cite{Griffith19, penny22}, essentially imposing these invariant measures by proxy. Here, we imposed the invariant measures as constraints directly in the training algorithm, allowing the RC to generalize with fewer data. Fortunately, we have found that much of the value of this additional constraint is achieved through the use of the leading Lyapunov exponent. While the entire Lyapunov spectrum can be quite difficult to calculate, particularly for large systems, the leading Lyapunov exponent can be estimated by using numerical techniques such as the breeding method \cite{Toth93, Toth97} or other methods described in \cite{Ispolatov15, Geist90, Benettin76}. This provides an opportunity for extension of this technique to higher-dimensional systems.

Recent works from \cite{pathak22, lam22, bi22} have shown promise in producing data-driven surrogate weather models that are competitive by some metrics with conventional operational forecast models. A key property that has not yet been demonstrated with such surrogate models is their ability to reproduce dynamical quantities such as the LEs, which indicate an average measure of the response to small errors in the initial conditions. For weather models in particular, the enforcement of LEs is crucial for the correct operation of data assimilation algorithms \cite{kalnay2002}. Platt et al. \cite{platt21} demonstrated the importance of reconstructing the LE spectrum for producing a skillful deterministic forecast model. Similarly, Penny et. al. \cite{penny22} indicated the ability of the RC to reproduce accurate finite-time LEs as a requirement for RC-based ensemble forecasts to produce good estimates of the forecast error covariance, which is the primary tool used in conventional data assimilation methods to project observational data to unobserved components of the system. This information can then be used to make the RC robust to sparse and noisy measurements. While this is the more realistic scenario used in online weather prediction systems, it is a fact that is rarely taken into account in neural network applications. The introduction of explicit constraints in the training cost function both improves prediction and trains the RNN to reconstruct the correctly shaped attractor \cite{Lu18, platt22}.

\section{Acknowledgements}
J.A. Platt, S.G. Penny, and H.D.I. Abarbanel acknowledge support from the Office of Naval Research (ONR) grants N00014-19-1-2522 and N00014-20-1-2580. S.G. Penny and T.A. Smith acknowledge support from NOAA grant NA20OAR4600277. T.-C. Chen acknowledges support from the NOAA Cooperative Agreement with the Cooperative Institute for Research in Environmental Sciences at the University of Colorado Boulder, NA17OAR4320101.

\section{Source Code}
The basic RC implementation used in this study is available \\\url{https://github.com/japlatt/BasicReservoirComputing}

\printbibliography

\end{document}